\documentclass[12pt]{article}

\usepackage{amssymb,amsmath,amsfonts,eurosym,geometry,ulem,graphicx,caption,color,setspace,sectsty,comment,footmisc,natbib,pdflscape,subfigure,array,hyperref,enumitem,appendix,url}

\newtheorem{proposition}{Proposition}

\newtheorem{definition}{Definition}

\newcommand{\vect}{\boldsymbol}

\newcolumntype{L}[1]{>{\raggedright\let\newline\\arraybackslash\hspace{0pt}}m{#1}}
\newcolumntype{C}[1]{>{\centering\let\newline\\arraybackslash\hspace{0pt}}m{#1}}
\newcolumntype{R}[1]{>{\raggedleft\let\newline\\arraybackslash\hspace{0pt}}m{#1}}

\hypersetup{colorlinks=true,allcolors=blue}

\begin{document}

\hypersetup{pageanchor=false}
\begin{titlepage}
\title{A Theory of Multilevel Interactive Equilibrium in NeuroAI}
\author{Zhe Sage Chen\thanks{Department of Psychiatry, New York University Grossman School of Medicine, New York, NY 10016, USA; Department of Neuroscience, New York University Grossman School of Medicine, New York, NY 10016, USA; Institute for Translational Neuroscience, New York University Grossman School of Medicine, New York, NY 10016, USA; Department of Biomedical Engineering, Tandon School of Engineering, New York University, Brooklyn, NY 11201, USA. Email: zhe.chen@nyulangone.org}
\and Quanyan Zhu\thanks{Department of Electrical and Computer Engineering, Tandon School of Engineering, New York University, Brooklyn, NY 11201, USA. Email: qz494@nyu.edu}}
\date{\today}
\maketitle

\begin{abstract}
\noindent We propose a game-theoretic framework for adaptive multi-agent intelligent systems. Unlike classical game theory, which often treats strategies as primitive objects chosen by perfectly rational agents, the proposed framework provides a mathematical foundation for studying equilibrium in NeuroAI and can be viewed as an extension of game theory under relaxed assumptions, including partial observability, bounded computation, and uncertainty. At its core, Multilevel Interactive Equilibrium (MIE) generalizes the classical Nash equilibrium to intelligent systems with internal computation. Rather than being defined solely at the level of observable behavior, equilibrium emerges when neural learning dynamics, cognitive representations, and behavioral strategies mutually stabilize between interacting agents. This framework applies uniformly to interactions between two biological brains, two artificial agents, or hybrid human-AI systems. We discuss applications of multilevel game theory to human-autonomous vehicle driving, human-machine interaction, human-large language model (LLM) interaction, and computational psychiatry. We also outline experimental strategies and computational methods for estimating MIE and discuss challenges and prospects for future research.\\
\vspace{0in}
\vspace{0in}

\bigskip
\end{abstract}
\setcounter{page}{0}
\thispagestyle{empty}
\end{titlepage}
\pagebreak \newpage
\hypersetup{pageanchor=true}

\doublespacing

\section{Introduction} \label{sec:introduction}

For much of its history, the science of intelligence—both biological and artificial—has been framed around a deceptively simple question: how does an agent optimize its behavior in a complex environment? Classical reinforcement learning (RL) and decision theory have largely approached this problem from a single-agent perspective, treating the environment as passive or stationary. But the ecological reality in which the brain evolved is fundamentally different. Organisms rarely act alone. Many real-world behaviors—from economic markets and social cooperation to conflict and negotiation—depend on the ability to anticipate and respond to other agents. In both cooperative and competitive environments, organisms operate within social and adversarial ecosystems, where outcomes depend not only on the environment but also on the actions, intentions, and strategies of other agents. In such settings, behavior becomes intrinsically interactive, and game theory has offered a normative framework for reasoning about strategic interactions in economics, political science, evolutionary biology \citep{GameTheory47}, and computational psychiatry \citep{CP12}. Modern advances in artificial intelligence (AI), LLMs, and robotics create an increasing number of human-AI and NeuroAI scenarios, where interacting agents evolve under uncertainty \citep{CIbook18}. Despite some efforts \citep{Razakatinana20,Ismail24}, to our knowledge, a unifying theory for characterizing NeuroAI applications across behavioral, cognitive, and neural levels is still lacking.

At the core of game theory, the Nash equilibrium describes a configuration of strategies in which no agent can unilaterally improve its payoff. Classical equilibrium theory assumes agents with unbounded rationality, complete knowledge of the game structure, and instantaneous access to optimal strategies. However, biological brains do not satisfy any of these assumptions. Neural systems are constrained by finite computation, partial observability, noisy information, and limited memory, and must learn strategies gradually through experience, continually adapting to the behaviors of other agents who are themselves learning and adapting. Recently, a notion of {\it neuronic Nash equilibrium} was proposed in the EEG-derived neural representation space \citep{Zhu25}, generalizing the classic game-theoretic framework for brain-machine interface (BMI)-enabled multi-agent behaviors and extending the notion of inter-individual neural synchronization \citep{Kelso13}. Additionally, game-theoretic equilibrium convergence has been reported in human-machine interactions, generating insight into the human-in-the-loop optimization paradigm for assistive devices
\citep{Chasnov25}.

Decision neuroscience and neuroeconomics investigate how the brain supports strategic interaction \citep{Glimcher14,Camerer23,Camerer09}, leading to Bayesian Theory of Mind (ToM) \citep{Yoshida08,Yoshida10}. Experimental paradigms derived from economic games have revealed neural circuits involved in reasoning about others’ actions and intentions. In competitive and cooperative environments, the brain encodes variables relevant to social decision-making, including expected rewards, prediction errors, and beliefs about opponents, suggesting that the brain represents not only the value of actions but also the anticipated strategies of other agents, where successful behavior requires predicting how others will respond. Parallel developments have taken place in AI. Multi-agent RL (MARL) theory and algorithms propose that learning may occur through repeated interaction between agents that continually adapt to each other \citep{multiRL,Lowe17,Li22,LiZhu26,Su26}, achieving remarkable successes in domains such as games and robotics \citep{Silver16}. Therefore, the study of intelligence—whether biological or artificial—confronts the problem of learning and acting within strategic environments populated by other adaptive agents.

Despite these advances, a theoretical gap remains. First, classical game theory treats equilibrium purely at the level of observable strategies and payoffs. However, in biological systems, the strategies that appear at the behavioral level are generated by deeper computational processes within the brain. Equilibrium should not be viewed as a single-layer concept but as a multilevel phenomenon emerging from the interaction of neural, cognitive, and behavioral processes. Second, traditional game theory and equilibrium theory often assume that social interactions occur between two human players or between AI agents, yet increasingly common human-AI interactions create the need for a new theoretical framework. Related game-theoretic perspectives on socio-technical networks, resilience, distributed systems, and cognitive security further motivate treating equilibrium as an interaction between strategic behavior, information structure, and adaptive control \citep{ZhuBasar25,ZhuBasar24,Abdallah25,HuangZhu23}. 

\begin{figure}[t]
\centering
\includegraphics[width=\textwidth]{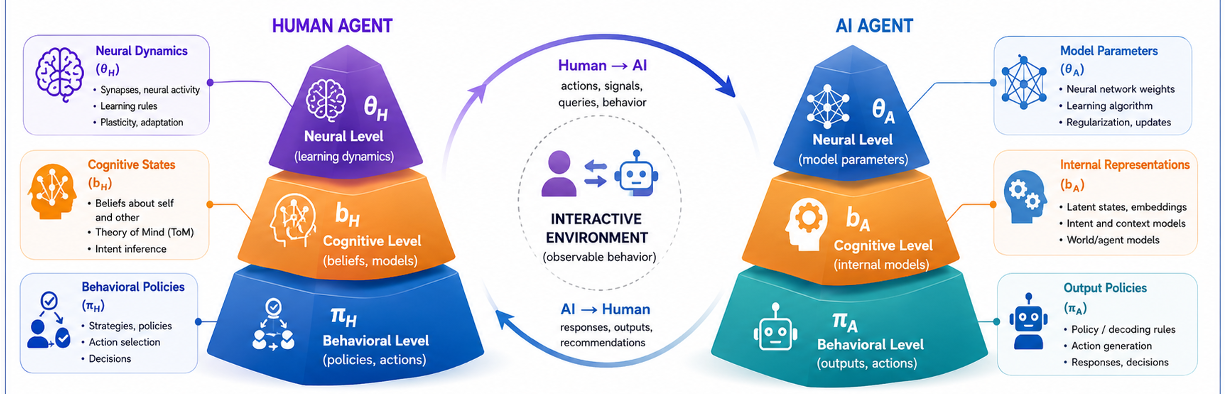}
\caption{\small Overview of the proposed NeuroAI game-theoretic framework and conceptual multilevel representation of Multilevel Interactive Equilibrium (MIE). The figure highlights how multilevel interaction couples internal learning processes, cognitive representations, behavioral strategies, and environmental feedback across biological, artificial, or hybrid agents. Human and artificial agents can be viewed as hierarchical systems consisting of neural, cognitive, and behavioral layers. Interaction occurs primarily at the behavioral level through the shared environment, while cognitive belief models and neural learning processes adapt through feedback across agents. Stable interaction emerges when neural learning, cognitive inference, and behavioral policies mutually stabilize, forming an MIE.}
\label{fig:framework}
\label{fig1}
\end{figure}

To address the gap, here we present a hierarchical multilevel equilibrium framework in which strategic stability emerges from interactions both within levels and across levels of the system. Within this framework, we envision that systems of intelligence, both biological and artificial, can achieve equilibrium at three levels: behavioral, cognitive, and neural levels, forming a conceptual multilevel structure (Fig.~\ref{fig1}).

\section{Multilevel Equilibrium Framework} \label{sec:framework}

In the hierarchy, at the base lies the {\it neural level}, where distributed populations of neurons encode value signals, prediction errors, and internal models of other agents. Neural activity implements learning rules—such as RL, active inference \citep{ActiveInference}, predictive coding, or synaptic plasticity—that update beliefs and policies during interaction. Above this sits the {\it cognitive level}, which includes internal representations such as beliefs, intentions, expectations, and models of opponents. At this level, agents maintain structured hypotheses about other agents' strategies and update them through inference. Finally, at the top of the pyramid lies the {\it behavioral level}, where actions, choices, and policies are expressed and observed. It is at this level that classical game-theoretic concepts: strategies, payoffs, and equilibria, are typically defined.

Within this framework, equilibrium can emerge in several different forms. First, within-level equilibria occur when stability is reached inside a single layer of the system. For example, at the behavioral level, two agents may converge to a stable pattern of strategies resembling a Nash equilibrium in a repeated game. At the cognitive level, equilibrium may correspond to a stable configuration of beliefs about the opponent’s strategy. At the neural level, equilibrium may appear as stable attractor dynamics in neural populations that represent expected value or opponent models.

Second, equilibria may occur across levels of organization. In biological systems, stable behavior often requires alignment between neural dynamics, cognitive representations, and overt strategies. For example, a behavioral equilibrium in a repeated game may arise only when neural learning signals stabilize the internal beliefs that support the strategy. In this sense, equilibrium can be viewed as a cross-level consistency condition: neural activity generates cognitive models that support behavioral strategies that in turn stabilize neural prediction errors.

More importantly, this hierarchical framework also allows equilibrium to be defined across different types of interacting agents. Traditional game theory primarily considers social or economic interactions among human decision-makers, but developments in AI and NeuroAI have expanded the scope considerably. Equilibrium may occur between two artificial agents engaging in self-play or competitive learning environments, and it can also emerge between a biological brain and an artificial agent, such as in human-AI collaboration or BMIs with artificial copilots \citep{Lee25}. In such hybrid systems, stability depends on whether both agents can successfully model and adapt to each other’s behavior across multiple levels of representation.

From this perspective, equilibrium is no longer merely a mathematical property of strategic games. Instead, it becomes an emergent property of interacting intelligent systems, arising from coupled dynamics across neural, cognitive, and behavioral layers. The pyramid representation helps clarify how these levels interact: neural mechanisms support cognitive models, cognitive models guide behavior, and behavioral outcomes are fed back to update neural representations. When these processes mutually stabilize—either within a single agent or between interacting agents—an equilibrium state may emerge.

\section{Theory} \label{sec:theory}

Here we propose a theory of multilevel equilibrium of interactive intelligence, where equilibrium is not merely a property of observable strategies but an emergent condition arising from coupled dynamics across neural, cognitive, and behavioral layers of interacting agents. These layers may operate on different timescales: neural or algorithmic adaptation can be slow, belief updating can be intermediate, and behavioral responses can be fast. The equilibrium is a fixed point or invariant set of that coupled dynamical system, from which we derive the within-level equilibrium (fixed point in one subsystem), cross-level equilibrium (simultaneous consistency across subsystems), and cross-agent equilibrium (coupling between agents' states). This conceptual framing aligns naturally with NeuroAI because both biological and artificial systems learn, predict, and adapt through hierarchical representations.

\subsection*{Setting}

Let there be $N$ agents $i\in \{1,\cdots,N\}$. Each time $t=0,1,2,\cdots$:
\begin{itemize}
    \item 	Environment state: $s_t \in \mathcal{S}$;
	\item  Action of agent $i$: $a_t^i\in \mathcal{A}_i$;
	\item Joint action: $\vect a_t=(a_t^1,\cdots,a_t^N)\in \mathcal{A}=\prod_i \mathcal{A}_i$; 
    \item Reward (or utility) to agent $i$: $r_i (s_t,\vect a_t)\in\mathbb{R}$;
    \item State transition: $s_{t+1}\sim P(\cdot | s_t,\vect a_t)$.
\end{itemize}
This is a Markov game (a.k.a. stochastic game) \citep{Littman94}.

\subsection*{The Three-Level Pyramid as a State Space}

To formalize the multilevel structure of intelligent agents, we represent each agent as a hierarchical system consisting of neural, cognitive, and behavioral states. These layers
form a conceptual pyramid in which neural mechanisms support cognitive representations, which in turn generate behavioral strategies.

\paragraph*{Neural level (substrate of learning)}
At the base of the pyramid lies the neural or algorithmic substrate that implements learning and inference. For agent $i$, we denote the neural state by
$\theta_t^i \in \Theta_i$,
where $\theta_t^i$ represents neural parameters or internal weights that govern learning
dynamics (e.g., synaptic strengths in biological brains or network parameters in
artificial systems).

\paragraph*{Cognitive level (beliefs and internal models)}
The intermediate layer represents internal cognitive states that encode beliefs about the
environment and other agents. We denote the cognitive state by $
b_t^i \in \mathcal{B}_i$,
where $b_t^i$ represents probabilistic beliefs, latent representations of opponent
strategies, or internal models of the environment.

\paragraph*{Behavioral level (policies or strategies)}
At the top of the pyramid lies observable behaviors. Agent $i$ selects actions according
to a policy $\pi_t^i \in \Pi_i$,
where $\pi_t^i$ maps environmental states $s_t$ and internal beliefs $b_t^i$ to action distributions.

\paragraph*{Multilevel agent state}
The complete internal state of agent $i$ is given by the tuple
$\vect x_t^i = (\theta_t^i, b_t^i, \pi_t^i)$,
which belongs to the product space
\[
\mathcal{X}_i = \Theta_i \times \mathcal{B}_i \times \Pi_i.
\]
For a system with $N$ interacting agents, the joint multilevel state is
\[
\vect x_t = (\vect x_t^1, \vect x_t^2, \dots, \vect x_t^N) \in \mathcal{X},
\]
where $\mathcal{X} = \prod_{i=1}^{N} \mathcal{X}_i$.
This representation defines a hierarchical state space in which neural learning,
cognitive inference, and behavioral strategies co-evolve during interaction.

This notation is intentionally broad. In empirical studies, $\pi_t^i$ is often partially observable through behavior, $b_t^i$ must usually be inferred as a latent cognitive state, and $\theta_t^i$ may correspond either to directly measured neural variables or to hidden algorithmic parameters. The framework therefore separates the conceptual levels of the agent from the measurement model used to estimate them.

\begin{figure}[t]
\centering
\includegraphics[width=0.82\textwidth]{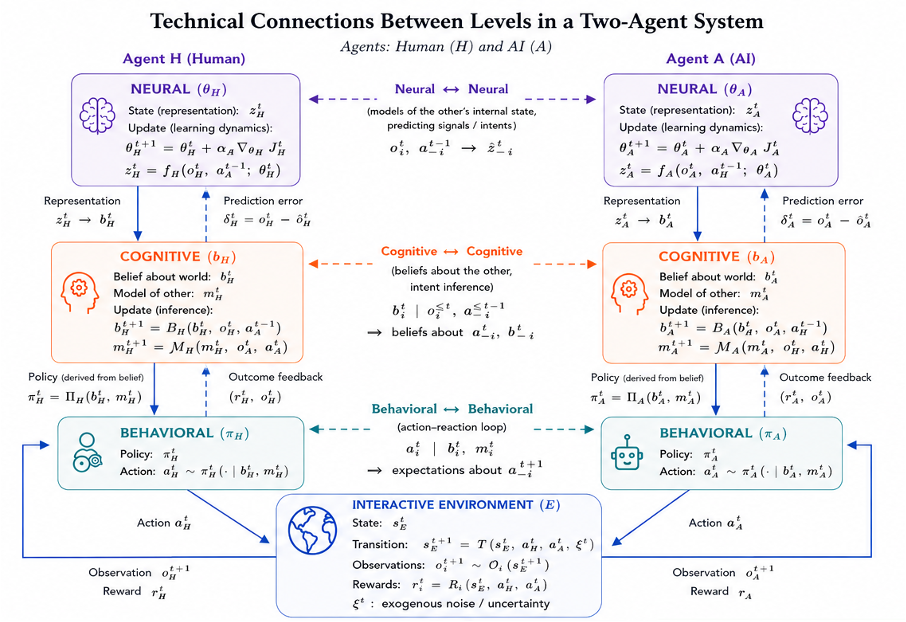}
\caption{\small Mathematical organization of the multilevel state representation. Each agent is described by neural or algorithmic variables $\theta$, cognitive belief states $b$, and behavioral policies $\pi$. The update operators couple these levels within each agent, while the environment couples agents through observations, actions, and rewards.}
\label{fig:mathlevel}
\end{figure}

\subsection*{Coupled Dynamics Across Levels}

The multilevel state representation allows us to formalize how neural learning,
cognitive inference, and behavioral strategies evolve during interaction.
These processes form a set of coupled dynamical systems that operate both
within each agent and across interacting agents.

\paragraph*{Action generation}
At time $t$, agent $i$ produces an action according to its behavioral policy.
Conditioned on the environmental state $s_t$ and cognitive belief state $b_t^i$,
\[
a_t^i \sim \pi_t^i(\cdot \mid s_t, b_t^i).
\]
The joint action of all agents is $
\vect a_t = (a_t^1, a_t^2, \dots, a_t^N)$. 
The environment then evolves according to the state transition dynamics $
s_{t+1} \sim P(\cdot \mid s_t, a_t)$.

\paragraph*{Cognitive update}
Agents update their internal models of the environment and other agents
based on observations. Let $o_t^i$ denote the observation available to
agent $i$ at time $t$; we assume that belief updates follow an operator
\[
b_{t+1}^i = F_i(b_t^i, o_t^i),
\]
where $F_i$ may represent Bayesian belief updating, inference over
latent opponent strategies, or neural network-based intent estimation.

In Bayesian form, belief update can be written as \citep{Khalvati19}
\[
b_{t+1}^i(\xi) \propto b_t^i(\xi)\, p(o_t^i \mid \xi),
\]
where $\xi$ denotes a latent variable such as an opponent type or
behavioral policy.

\paragraph*{Neural learning}
At the neural level, internal parameters adapt through learning signals
such as prediction errors or gradient updates. Let $\delta_t^i$ denote a
learning signal available to agent $i$. Neural parameters evolve as
\[
\theta_{t+1}^i = G_i(\theta_t^i, \delta_t^i).
\]
A common form is stochastic gradient learning
\[
\theta_{t+1}^i = \theta_t^i + \alpha_t
\nabla_{\theta^i} J_i(\theta_t^i ; \vect x_t^{-i}),
\]
where $\alpha_t$ is a learning rate, $J_i$ is the expected return
objective for agent $i$, and $\vect x_t^{-i}$ denotes the joint multilevel state excluding agent $i$.

\paragraph*{Policy adaptation}
Behavioral policies are generated from neural parameters and cognitive
belief states. We assume that policy updates are described by
\[
\pi_{t+1}^i = H_i(\pi_t^i, \theta_{t+1}^i, b_{t+1}^i).
\]
For example, a softmax policy may be defined as
\[
\pi_{t+1}^i(a \mid s, b)
\propto
\exp\!\left(\beta Q_{\theta_{t+1}^i}(s,a; b)\right),
\]
where $Q_{\theta}$ denotes a value function parameterized by neural
parameters $\theta$, and $\beta>0$ controls the degree of exploration.

\paragraph*{Global dynamical system}
Combining the environmental state and the multilevel agent states yields
a global system state
\[
\vect z_t = (s_t, \vect x_t) \in \mathcal{Z},
\]
where $\mathcal{Z} = \mathcal{S} \times \mathcal{X}$.

The evolution of the full system can be described by a Markov
process
\[
\vect z_{t+1} \sim \mathcal{T}(\cdot \mid \vect z_t),
\]
where $\mathcal{T}$ is the transition kernel induced jointly by the
environmental dynamics and the multilevel update operators
$\{F_i, G_i, H_i\}_{i=1}^N$.

Collectively, the above formulation makes explicit that neural learning, cognitive belief
updates, and behavioral policies evolve through a coupled dynamical
system across interacting agents.

\subsection*{Multilevel Interactive Equilibrium}

The multilevel dynamical system defined above allows us to formalize the
notion of equilibrium in interacting intelligent systems. Classical
game theory defines equilibrium purely at the level of behavioral
strategies. In contrast, our framework defines the equilibrium across
neural learning, cognitive inference, and behavioral policy dynamics.

\begin{definition}[Neural equilibrium]
A neural parameter profile $\theta^\ast = (\theta_1^\ast,\dots,\theta_N^\ast)$
is said to be a neural equilibrium if the expected neural learning dynamics
are stationary. Formally, for each agent $i$,
\[
\mathbb{E}\big[ G_i(\theta_i^\ast,\delta^i) \big] = \theta_i^\ast .
\]
For gradient-based learning rules, this condition is equivalent to
\[
\nabla_{\theta_i} J_i(\theta_i^\ast) = 0,
\]
where $J_i$ denotes the expected return for agent $i$.
\end{definition}

\begin{definition}[Cognitive equilibrium]
A belief profile $b^\ast = (b_1^\ast,\dots,b_N^\ast)$ is a cognitive
equilibrium if beliefs are invariant under the belief update operator
induced by interactions. That is, for each agent $i$,
\[
b_i^\ast = \mathbb{E}\left[ F_i(b_i^\ast, o^i) \right],
\]
where the expectation is taken with respect to the stationary
distribution of observations generated by the system.
\end{definition}

\begin{definition}[Behavioral equilibrium]
Given neural and cognitive states $(\theta,b)$, a policy profile
$\pi^\ast = (\pi_1^\ast,\dots,\pi_N^\ast)$ is a behavioral equilibrium
if no agent can improve its expected return through unilateral deviation.
Formally,
\[
\pi_i^\ast \in
\arg\max_{\pi_i \in \Pi_i}
V_i(\pi_i,\pi_{-i}^\ast;\theta,b),
\]
for every agent $i$, where $V_i$ denotes the expected value of agent $i$
under the joint policy profile.
\end{definition}
A special case of two-agent behavioral Nash equilibrium has been discussed earlier \citep{Zhu25b}.

\begin{definition}[Multilevel Interactive Equilibrium]
A tuple $(\theta^\ast, b^\ast, \pi^\ast)$
is a \textit{Multilevel Interactive Equilibrium (MIE)} if all three
conditions hold simultaneously:
\begin{enumerate}[label=\roman*.]
\item $\theta^\ast$ is a neural equilibrium (``neural stationarity");
\item $b^\ast$ is a cognitive equilibrium (``cognitive self-consistency");
\item $\pi^\ast$ is a behavioral equilibrium given $(\theta^\ast,b^\ast)$ (``behavioral best response").
\end{enumerate}
Equivalently, the multilevel state is a fixed point of the coupled update
operators that govern neural learning, cognitive inference, and behavioral
adaptation.
\end{definition}
The notion of MIE assumes three updates simultaneously reach equilibria; when the coupled systems are only partially observed (e.g., neural activity is unobserved), we refer to the system that satisfies two of three conditions (e.g., ii+iii) as a marginal MIE.

\begin{proposition}[MIE as a dynamical fixed-point]
Let $\Phi$ denote the joint update operator of the multilevel system,
$
\Phi(\theta,b,\pi)
=
\big\{
G(\theta,\delta),
F(b,o),
H(\pi,\theta',b')
\big\}
$; if 
\[
(\theta^\ast,b^\ast,\pi^\ast)
=
\mathbb{E}\big[\Phi(\theta^\ast,b^\ast,\pi^\ast)\big],
\]
then $(\theta^\ast,b^\ast,\pi^\ast)$ constitutes an MIE.
\end{proposition}

\paragraph*{Multilevel Equilibrium Principle}

In a Markov game where each agent updates (i) neural parameters by stochastic approximation, (ii) beliefs by consistent filtering, and (iii) policies by a smooth mapping from $(\theta, b)$, any limit point $(\theta^\star,b^\star,\pi^\star)$ of the coupled dynamics that is stationary under the induced kernel is an MIE; conversely, any stable MIE corresponds to an attracting invariant set of the multilevel learning dynamics. This framework unifies human-human, AI-AI, and human-AI interactions and distinguishes different cases:
\begin{itemize}
\item {\bf Within-level equilibrium:} fixed point of $F$ only (cognitive), $G$ only (neural), or best-response condition (behavior); this can apply to both human-human (e.g., $F_i$ approximates Bayesian ToM, $G_i$ embodies synaptic learning constraints) and AI-AI interactions (e.g., $F_i$ is an explicit opponent model, $G_i$ implements error backpropagation).
\item {\bf Cross-level equilibrium:} MIE fixed point of the coupled operator.
\item	{\bf Cross-agent equilibrium:} MIE applies to any mixture of humans/AI via heterogeneous operators.
\item	{\bf Failure modes:} when no MIE exists, or when MIE exists but is unstable.
\end{itemize}

MIE generalizes classical Nash equilibrium
to intelligent systems with internal computation. Rather than being defined
solely at the level of observable strategies, equilibrium arises when
neural learning dynamics, cognitive belief models, and behavioral policies
mutually stabilize across interacting agents.

\subsection*{Stability and and Equilibrium Selection}

The definition of MIE characterizes
equilibrium as a fixed point of coupled multilevel dynamics.
In many interactive systems, however, multiple equilibria may exist.
Understanding which equilibria are reached requires analyzing the
\emph{stability} of these fixed points under the learning and inference dynamics.

\paragraph*{Multilevel dynamical system}
Let $\vect x_t = (\theta_t, b_t, \pi_t)$
denote the global multilevel state of the system; the evolution of the system can be written as
$\vect x_{t+1} = \Phi(\vect x_t)$,
where $\Phi$ denotes the joint update operator induced by  neural learning dynamics, belief updates, and policy adaptation.
An MIE $\vect x^\ast$ is said to be
\emph{locally asymptotically stable} if there exists a neighborhood
$\mathcal{U}$ of $\vect x^\ast$ such that for any initial condition
$\vect x_0 \in \mathcal{U}$,
the dynamical system converges to equilibrium:
$\lim_{t \to \infty} \vect x_t =\vect x^\ast$.
In stochastic learning systems, convergence may be defined in
probability or almost surely.

\begin{definition}[Local stability of MIE]
Consider the mean-field approximation of the dynamics
\[
\vect x_{t+1} = \bar{\Phi}(\vect x_t),
\]
where $\bar{\Phi}$ denotes the expected update operator.
Let ${\bf J} = D\bar{\Phi}(y^\ast)$
be the Jacobian matrix of the differential (or update) operator evaluated at 
equilibrium. A sufficient condition for local stability is that the spectral
radius of the Jacobian satisfies
\[
\rho({\bf J}) < 1,
\]
where $\rho(\cdot)$ denotes the largest magnitude eigenvalue.
\end{definition}

In the multilevel framework, stability depends jointly on neural
learning rates, belief inference dynamics, and behavioral policy
adaptation. Instabilities may arise when these processes operate on
incompatible time scales or when feedback loops between agents amplify
prediction errors. In contrast, a stable interaction emerges when
multilevel learning dynamics converges to a common attractor in the
joint state space.

\paragraph*{Equilibrium selection}

When multiple equilibria exist, the equilibrium reached by the
system depends on initial conditions and the geometry of the
learning dynamics. Each equilibrium corresponds to an
\emph{attractor basin}: 
\[
\mathcal{B}(\vect x^\ast) =
\{\vect x_0 : \lim_{t\to\infty} \vect x_t =\vect x^\ast \}.
\]
Different equilibria may correspond to different modes of interaction, including cooperative, competitive, or unstable regimes.

\section{Applications of Multilevel Game Theory} \label{sec:applications}

\subsection*{Human-Autonomous Vehicle Interaction as a Multilevel Game}

Autonomous driving offers an excellent illustration of the multilevel equilibrium theory because the environment involves continuous strategic interaction
between human drivers and artificial agents controlling autonomous
vehicles \citep{Shu23,LiTraffic25,ZhaoShared24,LeiDriving23}. Human drivers continuously predict the actions of others, while autonomous systems must infer human intent and adapt their control policies accordingly. Therefore, a vehicle–driver interaction forms a two-agent stochastic dynamic game, in which equilibrium corresponds to stable and predictable driving behavior emerging from coupled human cognition and machine decision-making.

\paragraph*{Traffic dynamics}
Consider a human driver $H$ interacting with an autonomous vehicle $A$.
Let $s_t \in \mathcal{S}$ denote the traffic state at time $t$,
which may include positions, velocities, road geometry, and other
relevant environmental variables. Each agent selects an action $
a_t^H \in \mathcal{A}_H, a_t^A \in \mathcal{A}_A$,
corresponding to control variables such as steering, acceleration,
or braking. The joint action is $
\vect a_t = (a_t^H, a_t^A)$.
The traffic state evolves according to
$s_{t+1} \sim P(\cdot \mid s_t, a_t^H, a_t^A)$.

\paragraph*{Multilevel representation of the agents}

Each agent is represented using the multilevel state formulation
$\vect 
x_t^i = (\theta_t^i, b_t^i, \pi_t^i)$ for $i \in \{H,A\}$.

\textit{Neural level.}
For the human driver, $\theta_t^H$ represents neural processes related to
perception, decision making, and motor control. For the autonomous vehicle,
$\theta_t^A$ represents parameters of the perception and control
algorithms (e.g., neural network weights used in trajectory prediction).

\textit{Cognitive level.}
Both agents maintain internal models of each other’s behavior. The human
driver forms beliefs about the autonomous vehicle
$b_t^H \approx P(a^A \mid s)$,
while the autonomous vehicle maintains a model of human intent
$b_t^A \approx P(a^H \mid s)$.

\textit{Behavioral level.}
Each agent selects actions according to the policy
\[
a_t^H \sim \pi_t^H(s_t, b_t^H), \qquad
a_t^A \sim \pi_t^A(s_t, b_t^A).
\]
These policies determine observable driving behavior.

\paragraph*{Multilevel interactive equilibrium in traffic}
A human-autonomous vehicle interaction reaches an
MIE when the tuple
$(\theta_H^\ast, \theta_A^\ast,
b_H^\ast, b_A^\ast,
\pi_H^\ast, \pi_A^\ast)
$ satisfies the equilibrium conditions: (i) 
the neural learning dynamics stabilizes for both agents:
$
\nabla_{\theta_H} J_H = 0,
\qquad
\nabla_{\theta_A} J_A = 0 $; (ii) both agents develop consistent expectations about each other’s behavior: $
b_H^\ast \approx P(a^A \mid s),
\qquad
b_A^\ast \approx P(a^H \mid s)$; (iii) 
the two driving strategies $(\pi_H^\ast, \pi_A^\ast)$ become mutually stable.

\paragraph*{Example}

Consider a highway merging scenario where a human driver attempts to
merge into a lane occupied by an autonomous vehicle. Initially, both
agents are uncertain about each other's behavior. Over repeated
interaction, the human driver learns that the autonomous vehicle tends
to yield, while the autonomous system learns that the human merges once
sufficient space becomes available. The interaction may converge to a
stable behavioral pattern: the autonomous vehicle slows slightly
to allow the merge while the human driver merges once the gap appears.
This configuration represents a multilevel equilibrium supported by
consistent neural expectations, cognitive belief models, and behavioral
policies.

\subsection*{Brain–Machine Interaction as a Two-Agent Game}

BMIs can be naturally modeled as an
interactive system involving two adaptive agents: a biological brain
and an artificial decoding system. Within the multilevel game framework,
the brain and machine jointly form a coupled dynamical system whose
behavior emerges from the interaction between neural activity,
decoder control, and task dynamics.

The BMI system consists of a biological brain $H$ and an artificial agent $M$ (e.g., a machine decoder).
The internal state of the brain is represented by
$\vect x_t^H = (\theta_t^H, b_t^H, \pi_t^H)$,
where
\begin{itemize}
\item $\theta_t^H$ denotes neural circuit parameters or plasticity states,
\item $b_t^H$ represents internal beliefs about the decoder mapping,
\item $\pi_t^H$ represents the neural policy that generates neural activity patterns $n_t$: $n_t \sim \pi_t^H(s_t, b_t^H)$.
\end{itemize}

The artificial decoder (such as a Kalman filter or a neural network) is represented by
$\vect x_t^M = (\theta_t^M, b_t^M, \pi_t^M)$,
where
\begin{itemize}
\item $\theta_t^M$ denotes decoder parameters,
\item $b_t^M$ represents the machine's model of user intent,
\item $\pi_t^M$ denotes the control policy that maps neural activity
to device commands $u_t$: $u_t \sim \pi_t^M(n_t, b_t^M)$.
\end{itemize}

Under this formulation, brain-machine interaction can be viewed as a
two-agent multilevel game in which neural learning, belief inference,
and behavioral control policies co-evolve.
Successful BMI control corresponds to a multilevel equilibrium between a biological brain and an artificial agent. Stable BMI control emerges when the coupled dynamics between the brain and the decoder converge
toward an MIE; namely, three processes converge simultaneously:
(i) neural plasticity in the brain; (ii) learning in the machine decoder; and (iii) mutual prediction between the brain and machine.
On the other hand, BMI instability implies the failure of equilibrium, when either neural adaptation and decoder learning operate on incompatible timescales, or belief models fail to converge. 
This can lead to nonstationary co-adaptation. Furthermore, decoder adaptation should respect the neural learning timescales.

The MIE framework also predicts stability if $\alpha_M\ll \alpha_H$, which matches empirical BMI findings where slow decoder updates stabilize learning \citep{Oweiss15}. 
As another prediction, shared internal models may improve equilibrium. For example, if the machine explicitly models the intent of the user 
$b_M\approx b_H$, the equilibrium may be reached faster.
This motivates intent-aware decoders and co-adaptive BMI algorithms \citep{Madduri21,Madduri24}.

\subsection*{Human-LLM Interaction }

Interactions between humans and LLMs such as
ChatGPT can be interpreted as a two-agent interactive
system. In this setting, a human user and a language model exchange
messages through a conversational interface. The dialog evolves as a
sequential decision process in which both agents adapt their responses
based on the evolving conversational context.

\paragraph*{Conversation dynamics}

Let $s_t \in \mathcal{S}$ denote the conversational state at time $t$, which can represent the dialog history, including previous prompts,
responses, and contextual information.

The human agent $H$ produces a prompt
$p_t \in \mathcal{P}$,
and the language model agent $M$ produces a response $
r_t \in \mathcal{R}$.
The conversational state evolves according to
\[
s_{t+1} = f(s_t, p_t, r_t),
\]
where $f$ denotes the dialog update operator that appends new messages
to the conversation context.

\paragraph*{Multilevel agent representation}

Within the multilevel framework, both agents are represented by neural,
cognitive, and behavioral states.

\textit{Human agent.}
The human internal state is represented as $
x_t^H = (\theta_t^H, b_t^H, \pi_t^H)$,
where
\begin{itemize}
\item $\theta_t^H$ represents the neural processes that underlie reasoning,
language production, and learning,
\item $b_t^H$ represents the user's internal belief about the LLM's
capabilities and behavior,
\item $\pi_t^H$ represents the prompting strategy used by the user.
\end{itemize}
The human generates prompts according to the policy 
\[
p_t \sim \pi_t^H(s_t, b_t^H).
\]

\textit{LLM agent.}
The language model agent is represented by
$x_t^M = (\theta_t^M, b_t^M, \pi_t^M)$,
where
\begin{itemize}
\item $\theta_t^M$ denotes the parameters of the language model,
\item $b_t^M$ represents the model's inferred representation of user
intent,
\item $\pi_t^M$ denotes the response generation policy.
\end{itemize}
Responses are generated according to the policy
\[
r_t \sim \pi_t^M(s_t, b_t^M).
\]

\paragraph*{Belief and learning updates}
Both agents update their internal representations through interaction.
The human updates expectations about the LLM based on observed responses,
\[
b_{t+1}^H = F_H(b_t^H, r_t),
\]
while the LLM updates its internal representation of user intent through
contextual inference,
\[
b_{t+1}^M = F_M(b_t^M, p_t).
\]
In addition, learning may occur across longer timescales through neural
adaptation in the human brain or through parameter updates in the LLM.

Therefore, human-LLM dialog can be viewed as a multilevel
interactive game in which conversational strategies emerge from the
coupled dynamics of neural cognition, belief inference, and language
generation policies. Efficient communication arises when the interaction
converges toward an MIE in which both the
human's prompting strategy and the model's response policy become
mutually predictable and stable. Prompt engineering reduces uncertainty in belief estimation and accelerates convergence, whereas failure of cognitive equilibrium generates misunderstanding or LLM hallucination. Preliminary theoretical analyses of behavioral equilibria in LLM-Nash games have been discussed in \citep{Zhu25b}; related Stackelberg, response-learning, and agentic-AI formulations provide complementary models for strategic LLM interaction and adaptation \citep{ZhuAI25,ZhuStackelberg25,XieZhu24,ZhuJailbreak25}.

\subsection*{Psychiatric Disorders as Disrupted Multilevel Equilibria}

In the multilevel framework, a healthy individual interacting with the environment can be described by the state $(\theta,b,\pi)$,
where $\theta$ represents neural parameters and circuit dynamics; $b$ represents cognitive beliefs or internal models of the world and other agents; $\pi$ represents behavioral policies that govern action. Healthy functioning corresponds to a stable multilevel equilibrium
$(\theta^\star,b^\star,\pi^\star)$,
in which neural learning, cognitive inference, and behavioral strategies remain mutually consistent with environmental feedback.
Psychiatric disorders can arise when this equilibrium is distorted, unstable, or trapped in maladaptive attractor states.

\paragraph*{Three Types of pathological equilibria}

According to the multilevel equilibrium framework, three types of pathological equilibria may occur. 

{\it Neural-level instability}: At the neural level, abnormal circuit dynamics or neurotransmitter signaling can destabilize learning processes. Examples of neural mechanisms include altered dopaminergic prediction errors, 	abnormal excitation–inhibition (E/I) balance, and 	impaired synaptic plasticity.
Formally, neural dynamics $G$ may not converge or may converge to pathological attractors.
This can propagate upward to cognitive and behavioral dysfunction.
 
{\it Cognitive-level miscalibration}:
At the cognitive level, disorders may involve incorrect belief updates about the environment or other agents.
If beliefs evolve as $b_{t+1}=F(b_t,o_t)$, pathology may arise when
$b^\star \not=\mathbb{E}[F(b^\star ,o)]$ 
or when beliefs converge to incorrect fixed points.
Examples of such behavior include exaggerated threat expectations,
	incorrect inference of social intent, and persistent negative self-beliefs.
 
{\it Behavioral-level maladaptive strategies}:
At the behavioral level, individuals can adopt strategies that are locally stable but globally maladaptive.
For example, $\pi^\star\not= \arg \max V(\pi)$
with respect to long-term well-being.
These policies may remain stable because they are reinforced by distorted beliefs or neural signals.

\paragraph*{Examples of psychiatric phenotypes}

Many psychiatric disorders can be interpreted or framed as pathological equilibria. 
\begin{itemize}

\item {\bf Depression}, which may correspond to a negative cognitive equilibrium. At the neural level, reduced dopaminergic reward-prediction signals may alter learning dynamics.
At the cognitive level, beliefs about future reward become pessimistic (i.e., $b^\star$(reward)$\ll$ true reward probability). At the behavioral level, individuals adopt withdrawal strategies; consequently, the system stabilizes in a low-reward equilibrium.

\item {\bf Anxiety disorder}, which may correspond to a threat-biased equilibrium. In this case, cognitive beliefs overestimate risk (i.e., $b^\star$(threat)$\gg$ true threat probability), or behavioral strategies become more avoidance-based. Although avoidance temporarily reduces perceived risk, it prevents corrective evidence, reinforcing the equilibrium.

\item {\bf Schizophrenia or psychosis}, which may involve unstable belief dynamics. At the cognitive level, aberrant prediction errors lead to excessive belief updates; this can generate	delusional beliefs,
hallucinations, or impaired ToM models.

\item {\bf Autism spectrum disorders}, which may involve altered equilibrium in social interactive games. Differences in social belief modeling may lead to mismatches between internal models of other agents and observed social behavior. As a result, this can make social interaction equilibria more difficult to stabilize.

\end{itemize}

\paragraph*{Implications for treatment in precision psychiatry}

The new framework also suggests that effective treatments work by perturbing the system across levels. For example, at the neural level, pharmacotherapy targets neural parameters $\theta$; at the cognitive level, psychotherapy targets belief update processes $b$; at the behavioral level, behavioral interventions modify policy $\pi$. Accordingly, successful treatment can push the system out of a pathological attractor toward a healthier equilibrium. Since the equilibrium state depends on individual neural and cognitive parameters, the framework naturally supports personalized treatment strategies.

\subsection*{Illustrative Simulation: Human-LLM Co-adaptation}

Consider a simple problem: a human repeatedly asks an LLM to summarize scientific papers. Initially, the human uses underspecified prompts such as ``Summarize this,'' while the LLM responds generically. After repeated interactions, the user learns to provide more structured prompts, such as ``Summarize the main contribution, key methods, and limitations in three bullet points.'' In parallel, the LLM infers the user's preferred output format from context. Over time, prompt specificity and model alignment converge, task utility increases, and the interaction stabilizes into a multilevel equilibrium characterized by efficient and predictable communication. Figure~\ref{fig2} presents a simple illustrated example (see Appendix for details).

\begin{figure}[t]
\centering
\includegraphics[width=\textwidth]{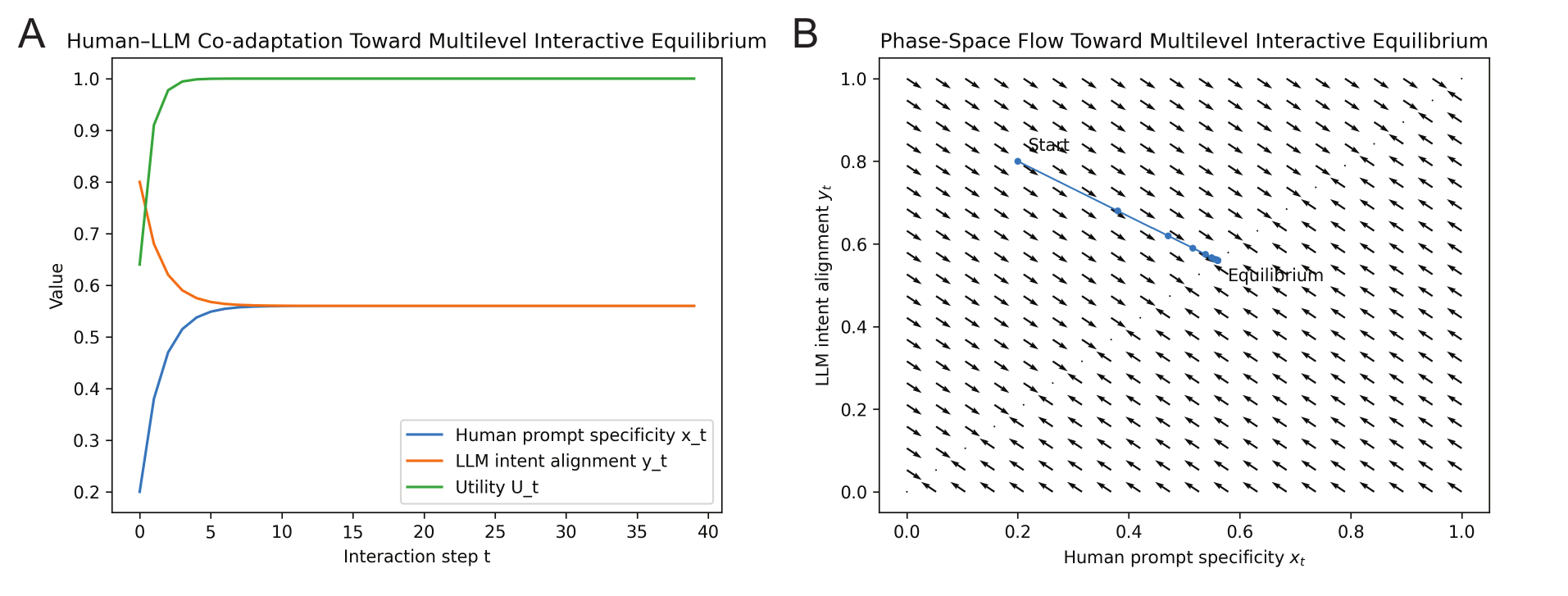}
\caption{\small A toy interaction model illustrates how prompt specificity $x_t$
 (human strategy) and intent alignment $y_t$
 (LLM inference) evolve through coupled adaptation dynamics. (A) Task utility $U_t$
increases as the mismatch between human prompting and model interpretation decreases. The system converges toward a fixed point or MIE where 
$x_t \approx y_t$. (B) The phase-space flow in $(x_t,y_t)$ space shows the coupled evolution of human prompt specificity $x_t$ and LLM intent alignment $y_t$. The blue trajectory shows convergence from an initial mismatch toward a stable fixed point along the diagonal.
}
\label{fig2}
\end{figure}

\section{Estimating MIE: Experimental Strategies and Computational Methods} \label{sec:estimation}

The MIE theory defines equilibrium
as a fixed point of coupled neural, cognitive, and behavioral dynamics. However, the theory does not directly provide a practical means to estimate the equilibrium. 

In practice, these quantities are rarely directly observable. Instead,
equilibrium must be estimated from empirical interaction data through a
combination of behavioral analysis, latent-variable inference, and
dynamical system modeling.

\paragraph*{Neural dynamics and stationarity}

At the neural level, equilibrium corresponds to stationary learning
dynamics. Neural signals $z_t$ (e.g., population activity or prediction
errors) can be analyzed using dynamical system models.
Neural equilibrium is supported 
 when prediction errors converge toward zero. Existing latent dynamical system techniques, such as factor analysis,
state-space modeling, or neural manifold learning, can be used to
identify attractor structures in neural population activity.

One approach is to analyze the latent state trajectory using attractor analysis. A fixed point, line attractor, limit cycle, or periodic oscillator may reveal stationary patterns of neural dynamics at different timescales. Another approach is neural decoding of belief and policies. In a BMI study, one can train decoders from neural activity to predict choice or beliefs about an opponent's action, and then test whether these decoded variables stabilize across repeated interaction. 

We can also jointly examine neural activities of two agents. One possible approach is to linearly project them separately into a low-dimensional subspace that maximizes their correlation (e.g., canonical correlation analysis or CCA), or maximize their covariance matrix by partial least-squared correlation (PLSC). Once the shared component in the subspace is identified, we can monitor the convergence of neural population trajectories of two agents projected separately in the dominant shared subspace.

\paragraph*{Cognitive state estimation}

Cognitive belief states are typically latent and must be inferred from
observed behavior. A common approach is to model beliefs as hidden
states in a state-space model. The multilevel
equilibrium is then estimated as a fixed point
of the inferred dynamical system.
Belief trajectories can then be estimated using Kalman filtering, approximate Bayesian filtering, or variational inference. Cognitive equilibrium is
suggested when inferred beliefs stabilize,
\[
\|\hat{b}_{t+1}^i - \hat{b}_t^i\| < \varepsilon .
\]
Experimental probes, such as eliciting explicit predictions about an
opponent's actions, can further validate inferred belief states.

One can also design experimental probes to reveal each agent's internal model of the other. For humans, this opponent-modeling strategy includes predicting the opponent's next move and collecting trial-by-trial confidence estimates. For AI systems, this strategy includes estimating the inferred user's intent, analyzing attention to the user history, and fitting interpretable surrogate models. The inferred beliefs $\hat{b}_i$ can then be compared with the observed opponent policy $\hat{\pi}_{-i}$ using Kullback-Leibler divergence, Wasserstein distance, or predictive log likelihood.

Another strategy involves recursive belief modeling or iterative reasoning. If agent $i$ represents beliefs about another agent's beliefs, first-order beliefs can be ``what I think you will do,'' and second-order beliefs can be ``what I think you think I will do,'' yielding a hierarchy of beliefs \citep{Zhu26}.
In human-LLM interaction or psychiatric social inference experiments, one can fit nested belief models such as $b_i^{(1)}=\textit{belief about other's action}$ and $b_i^{(2)}=\textit{belief about other's belief}$, then compare models of different depths using held-out prediction.  

\paragraph*{Behavioral estimation}

At the behavioral level, one can run repeated trials of the same interaction, such as closed-loop BMI control or iterative human-LLM prompting. Given an observed history of state-action frequencies,
an empirical policy estimate may be obtained as
\[
\hat{\pi}_i(a \mid s) =
\frac{N_i(s,a)}{\sum_{a'} N_i(s,a')},
\]
where $N_i(s,a)$ denotes the number of times agent $i$ selects action
$a$ in state $s$. Behavioral equilibrium is suggested when policy drift
becomes small over time for all agents over a sustained interaction window:
\[
\|\hat{\pi}_{i,t+1} - \hat{\pi}_{i,t}\| < \varepsilon.
\]
 
An additional diagnostic is the \emph{best-response gap}: given estimated opponent policy $\hat{\pi}_{-i}$, compute whether the observed policy is close to a best response:
\[
\mathrm{BRGap}_i =
\max_{\pi_i} V_i(\pi_i,\hat{\pi}_{-i}) -
V_i(\hat{\pi}_i,\hat{\pi}_{-i}),
\]
which measures the incentive for unilateral deviation. If BRGap is sufficiently small for all agents, behavior is near equilibrium.

For a given stochastic dynamical policy model, $\pi_{t+1}=f_\phi(\pi_t,s_t,o_t)$, one can also estimate its fixed points and examine its behavioral dynamics. 

\paragraph*{Joint estimation}
A principled strategy is to jointly model neural, cognitive, and
behavioral processes using hierarchical state-space models,
\[
\theta_{t+1} = G(\theta_t,\delta_t) + \xi_t^\theta,
\]
\[
b_{t+1} = F(b_t,o_t) + \xi_t^b,
\]
\[
a_t \sim \pi(\cdot \mid s_t,b_t,\theta_t).
\]
where $\xi^\theta$ and $\xi^b$ denote the additive noise terms for $\theta$ and $b$ system equations, respectively.
Such models can be fitted using expectation-maximization (EM), variational
Bayes, or particle Markov chain Monte Carlo (MCMC) methods. 

An empirical metric for the ``distance to equilibrium" is
\begin{eqnarray*}
\mathcal{E}_t&=&\lambda_\theta \|\theta_{t+1}-\theta_t\| 
+  \lambda_b \|b_{t+1}-b_t\| 
+ \lambda_\pi \|\pi_{t+1}-\pi_t\| \nonumber\\
&& + \lambda_{\mathrm{BRGap}}\sum_i \textrm{BRGap}_i,
\end{eqnarray*}
where $\lambda_\theta,\lambda_b,\lambda_\pi$ and $\lambda_{\mathrm{BRGap}}$ are nonnegative weights that encode the relative importance and measurement reliability of the neural, cognitive, behavioral, and game-theoretic terms. 
When $\mathcal{E}_t$ becomes small and stays small, the system is near multilevel equilibrium.

 \subsection*{Experimental validation through perturbation}

Equilibrium stability can be experimentally tested through controlled
perturbations. After an equilibrium is reached, perturbations
may be introduced by modifying environmental conditions, altering agent
policies, or injecting neural or cognitive perturbations. A stable MIE
is expected to exhibit recovery toward the same fixed point, whereas
instability may lead to transitions toward alternative attractors.

Since a true equilibrium should be resistant to small perturbations, local stability can be tested experimentally after apparent convergence. One may perturb one layer, such as by changing reward contingencies, altering AI response style, modifying a decoder mapping, injecting misleading social cues, or applying neural stimulation, and then test whether the system returns to the same equilibrium, moves to a new one, or becomes unstable.

Furthermore, manipulation of timescales, such as the adaptation rates (e.g., fast vs. slow; fixed vs. adaptive) of one agent or one level, may allow us to test the theory's prediction that equilibrium depends on compatible cross-level learning timescales.

Additionally, one can consider masking or distorting information about the opponent in social interactions, or feedback information in BMI systems, to test how belief estimation affects equilibrium. 

MIE can be empirically tested by computer simulations. 
 Within the MARL framework, two agents are implemented by two recurrent neural networks (RNNs) that independently model the model-based actor-critic RL. In a cooperative or competitive game, one can estimate the shared neural subspace of two RNNs' activities and perturb the subspace to probe the behavioral impact \citep{Zhang25,Jiang26}.

\section{Discussion: Toward a Multilevel Science of Interactive Intelligence} \label{sec:discussion}

The multilevel equilibrium framework may provide a unifying perspective across a wide range of NeuroAI systems. It suggests that the study of strategic interaction should incorporate not only behavioral outcomes but also the internal computational mechanisms that generate them. It also provides a common framework for comparing biological and artificial intelligence, since both can be analyzed in terms of interacting levels of representation and learning dynamics. In BMI, equilibrium may correspond to a stable alignment between neural encoding and decoder policies. In conversational AI, human-LLM dialog may converge toward efficient communication strategies as users and models learn each other's behaviors. In autonomous driving, interactions between human drivers and automated systems may stabilize into socially compatible traffic conventions. 
Finally, it highlights the possibility that instability or dysfunction in social behavior—such as those observed in certain psychiatric conditions—may arise from misalignment across levels of the system, where neural learning dynamics fail to support stable cognitive models or behavioral strategies.

An outstanding issue concerns the existence and uniqueness of multilevel equilibria. Classical game theory has established conditions under which Nash equilibria exist, but analogous results for multilevel dynamical systems remain largely unexplored. In particular, the joint state space may admit multiple attractors corresponding to distinct behavioral regimes. Determining when such equilibria exist, how many may occur, and whether they are stable under realistic learning dynamics represents an important theoretical challenge. Recent work on Bayesian equilibria in resilient multiagent systems points to one possible route for connecting information structures, belief consistency, and equilibrium analysis \citep{Nugraha26}. Another open problem involves timescale separation across levels. Neural learning, belief updating, and behavioral adaptation often operate on different temporal scales. Understanding how equilibrium emerges when these processes interact asynchronously may require new analytical tools from stochastic dynamical systems and MARL. Furthermore, decision-making or actions in social interactions may be driven by mutually independent cognitive and non-cognitive components. How to incorporate non-cognitive components (such as cultural norm, group influence, impulsive drive) into the game-theoretical framework remains unclear. 

A common problem in MARL is instability caused by non-stationarity. Strategies with opponent-learning awareness (LOLA) can mitigate this problem \citep{Foerster18}. Another promising approach is the embedded predictive intelligence approach \citep{Meulemans25}, in which embedded Bayesian agents model self and others and distinguish ``self-similar'' agents from ``decoupled'' agents, where interactions with self-similar agents can lead to a coherent framework for MARL. Applying this framework to estimate cognitive or behavioral equilibrium may prove fruitful for rational games.

In practice, testing the multilevel equilibrium hypothesis requires measurements that simultaneously capture neural activity, cognitive inference, and behavior. With advances in large-scale neural recording and computational modeling, a key methodological challenge is the identification of latent cognitive variables, such as internal beliefs or opponent models, from behavioral data.

Several directions may be particularly promising for future work.
First, development of computational tools for estimating MIE from real-world interaction data is critical. Advances in hierarchical state-space modeling and deep dynamical systems may enable joint estimation of neural, cognitive, and behavioral dynamics.
Second, theoretical work is needed to connect MIE with existing frameworks such as MARL and evolutionary game dynamics. Establishing these links could place the theory within a broader mathematical landscape.
Third, future research should explore collective multilevel equilibria in larger populations of interacting agents. Real-world systems such as social networks, traffic systems, human-robot and human-AI ecosystems often involve many agents whose interactions generate emergent group-level behavior. Fourth, digital twins and MARL frameworks can provide insight into the theory and estimation of MIE given complete neural observability \citep{Zhang25,LiTraffic25}; neural perturbation in silico can further causally test new hypotheses or suggest innovative human intervention strategies. Fifth, applying multi-agent game theory to psychiatry has gained growing interest \citep{Loula20,Cullen18}; combining game-behavior paradigms, cognitive tasks, and EEG/fMRI neuroimaging can open new avenues for probing social decision-making and social trust in psychiatric disorders \citep{Rilling11,Razin21}.

Ultimately, the concept of MIE suggests a shift in how we think about intelligent systems. Rather than viewing intelligence as the property of isolated agents optimizing static objectives, this framework emphasizes adaptive agents embedded in interactive environments, continuously learning and predicting one another across multiple levels of representation. As human and AI become increasingly intertwined, understanding these multilevel interaction dynamics may become a central problem for neuroscience, AI, and the science of complex systems.

\section{Conclusion} \label{sec:conclusion}

In summary, our proposed pyramid framework reframes equilibrium as a multiscale phenomenon of interactive intelligence, linking neural computation, cognitive inference, and observable behavior across both biological and artificial agents. This perspective opens the door to a richer theory of game dynamics in NeuroAI, where equilibrium is understood as a stable configuration emerging from coupled processes across levels and agents.
More broadly, the framework suggests that many human–AI systems should be designed not merely to optimize performance, but to reach stable interactive equilibria that humans can easily predict and trust.

\clearpage
\onehalfspacing
\section*{Appendix} \label{sec:appendix}
\addcontentsline{toc}{section}{Appendix}

For a simple illustration of human-LLM co-adaptation, let $x_t \in [0,1]$ denote the specificity of the human prompt at time $t$,
and let $y_t \in [0,1]$ denote the degree to which the LLM's internal
representation is aligned with the user's intended task.

Assume that the interaction utility is defined as 
\[
U_t = 1 - (x_t-y_t)^2,
\]
so that task performance is maximized when prompt specificity and model alignment match (Fig.~\ref{fig2}A).
The human updates prompt specificity in response to task performance:
\[
x_{t+1} = x_t + \alpha_H \frac{\partial U_t}{\partial x_t}
        = x_t - 2\alpha_H(x_t-y_t),
\]
where $\alpha_H > 0$ denotes the human adaptation rate.

In parallel, the LLM updates its inferred alignment state using the observed prompt:
\[
y_{t+1} = y_t + \alpha_M(x_t-y_t),
\]
where $\alpha_M > 0$ denotes the model adaptation rate.

Defining the mismatch term
\[
d_t = x_t-y_t,
\]
the coupled human-LLM dynamics can be rewritten as a one-dimensional contraction mapping
\[
d_{t+1} = (1-2\alpha_H-\alpha_M)d_t.
\]
Thus, in light of the contraction mapping theorem \citep{Hunter20}, the interaction converges to equilibrium (i.e., $x^\ast = y^\ast$) if  
\[
|1-2\alpha_H-\alpha_M| < 1.
\]
Thus, stable communication emerges as a fixed point of human-LLM co-adaptation (Fig.~\ref{fig2}B).

\subsection*{Acknowledgments}

Z.S.C. was partially supported by the US National Institutes of Health (NIH) grants RF1-DA056394, R01-MH139352, and P50-MH132642.

\singlespacing
\setlength\bibsep{0pt}

\end{document}